\pdfoutput=1

\documentclass[10pt,twocolumn,letterpaper]{article}

\usepackage{cvpr}
\usepackage{times}
\usepackage{epsfig}
\usepackage{graphicx}
\usepackage{amsmath}
\usepackage{amssymb}
\usepackage{graphicx}  
\usepackage{algorithm} 
\usepackage{algpseudocode} 
\usepackage[normalem]{ulem}
\usepackage{makecell}
\usepackage{rotating}
\usepackage{booktabs}
\usepackage{gensymb}
\usepackage[dvipsnames,svgnames,x11names]{xcolor}
\usepackage[pagebackref=true,breaklinks=true,letterpaper=true,colorlinks,bookmarks=false]{hyperref}
\usepackage[toc,page]{appendix}

\newcommand{\PreserveBackslash}[1]{\let\temp=\\#1\let\\=\temp}
\newcolumntype{C}[1]{>{\PreserveBackslash\centering}p{#1}}
\newcolumntype{R}[1]{>{\PreserveBackslash\raggedleft}p{#1}}
\newcolumntype{L}[1]{>{\PreserveBackslash\raggedright}p{#1}}

\cvprfinalcopy
\ifcvprfinal\fi
\begin{document}

\title{Neural RGB$\rightarrow$D Sensing: Depth and Uncertainty from a Video Camera}

\author{Chao Liu$^{1,2}$\thanks{The authors contributed on this work when they were at NVIDIA.}  
\qquad Jinwei Gu$^{1,3}$\footnotemark[1]
\qquad Kihwan Kim$^1$ 
\qquad Srinivasa Narasimhan$^2$ 
\qquad Jan Kautz$^1$\\
$^1$NVIDIA 
\qquad $^2$Carnegie Mellon University
\qquad $^3$SenseTime
}

\maketitle 
\begin{abstract}
Depth sensing is crucial for 3D reconstruction and scene understanding. 
Active depth sensors provide dense metric measurements, but often suffer from limitations such as restricted operating ranges, low spatial resolution, sensor interference, and high power consumption. 
In this paper, we propose a deep learning (DL) method to estimate per-pixel depth and its uncertainty continuously from a monocular video stream, with the goal of effectively turning an RGB camera into an RGB-D camera.
Unlike prior DL-based methods, we estimate a depth probability distribution for each pixel rather than a single depth value, leading to an estimate of a 3D depth probability volume for each input frame. 
These depth probability volumes are accumulated over time under a Bayesian filtering framework as more incoming frames are processed sequentially, which effectively reduces depth uncertainty and improves accuracy, robustness, and temporal stability. 
Compared to prior work, the proposed approach achieves more accurate and stable results, and generalizes better to new datasets.
Experimental results also show the output of our approach can be directly fed into classical RGB-D based 3D scanning methods for 3D scene reconstruction.
\end{abstract}

%INTRO 
%--------------------------------------
\vspace{-1em}
\section{Introduction}
\label{sec:intro}

\vspace{-.5em}
Depth sensing is crucial for 3D reconstruction~\cite{Newcombe11KinectFusion,Niessner2013Hashing,Whelan15rss} and scene understanding~\cite{guptaECCV14,qi2017frustum,song15SUNRGBD}. 
Active depth sensors (\eg, time of flight cameras~\cite{horaud16TOF, Remondino13TOF}, LiDAR~\cite{Christian2013ASO}) measure dense metric depth, but often have
limited operating range (\eg, indoor) and spatial resolution~\cite{chan08}, consume more power, and suffer from multi-path reflection and interference between sensors~\cite{Maimone12}. 
In contrast, estimating depth directly from image(s) solves these issues, but faces other long-standing challenges such as scale ambiguity and drift for monocular methods~\cite{Saxena:2008DRS6}, as well as the correspondence problem and high computational cost for stereo~\cite{Tippetts16} and multi-view methods~\cite{Seitz:2006}. 
\begin{figure}[t]
	\centering
    \includegraphics[width=.99\linewidth]{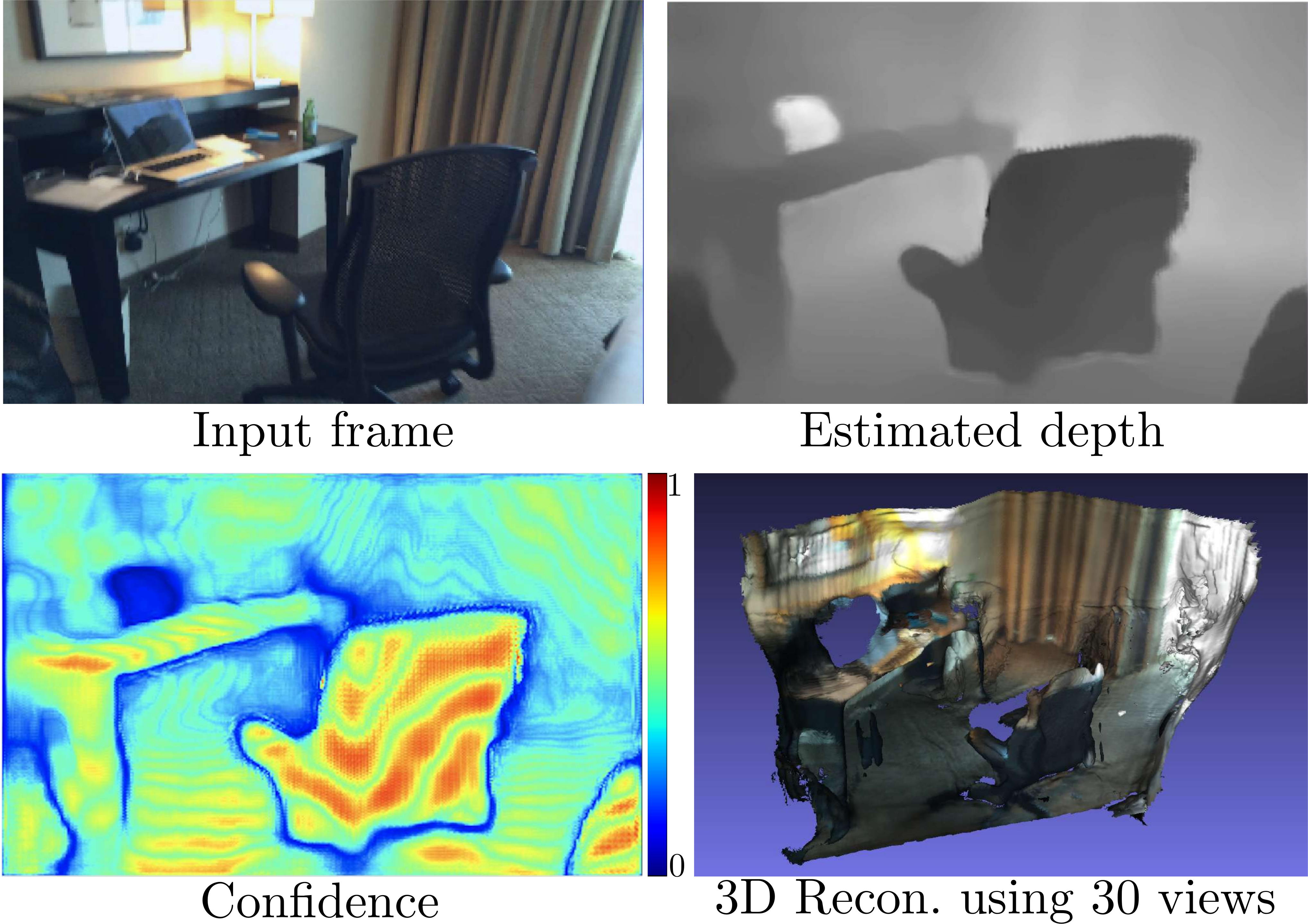}
\caption{
We proposed a DL-based method to estimate depth and its uncertainty (or, confidence) continuously for a monocular video stream, with the goal of turning an RGB camera into an RGB-D camera. 
Its output can be directly fed into classical RGB-D based 3D scanning methods~\cite{Newcombe11KinectFusion,Niessner2013Hashing}  
for 3D reconstruction.} 
	\label{fig:teaser}
	\vspace{-.5em}
\end{figure}   

Inspired by recent success of deep learning in 3D vision~\cite{Bloesch18CodeSLAM,Chang18PSM, Fu18DORN,Godard17MonoDepth,Huang18DeepMVS,Tateno17CNNSLAM, Ummenhofer17DeMoN,Wang18DDVO,Yao18MVSNet,Zhou18DeepTAM,Zhou17SfmLearner}, in this paper, we propose a DL-based method to estimate depth and its uncertainty continuously from a monocular video stream, with the goal of effectively turning an RGB camera into an RGB-D camera. 
We have two key ideas:
\begin{enumerate}
    \vspace{-.5em}
    \item Unlike prior work, for each pixel, we estimate a depth probability distribution rather than a single depth value, leading to an estimate of a Depth Probability Volume (DPV) for each input frame. 
    As shown in Fig.~\ref{fig:teaser}, the DPV provides both a Maximum-Likelihood-Estimate (MLE) of the depth map, as well as the corresponding per-pixel uncertainty measure. 
    
    \vspace{-.5em}
    \item These DPVs across different frames are accumulated over time, as more incoming frames are processed sequentially. %
    The accumulation step, originated from the Bayesian filtering theory and implemented as a learnable deep network, effectively reduces depth uncertainty and improves accuracy, robustness, and temporal stability over time, as shown later in Sec.~\ref{sec:results}. 
\end{enumerate}

\vspace{-.5em} 
We argue that all DL-based depth estimation methods should predict \emph{not depth values but depth distributions}, and should \emph{integrate such statistical distributions over time} (\eg, via Bayesian filtering).
This is because dense depth estimation from image(s) -- especially for single-view methods -- inherently has a lot of uncertainty, due to factors such as lack of texture, specular/transparent material, occlusion, and scale drift. 
While some recent work started focusing on uncertainty estimation~\cite{Gal2016Dropout,Ilg18Uncertainty,Kendall2017uncertainty,Kendall2018multi} for certain computer vision tasks, to our knowledge, we are the first to predict a depth probability volume from images and integrate it over time in a statistical framework. 

We evaluate our method extensively on multiple datasets and compare with recent state-of-the-art, DL-based, depth estimation methods~\cite{Fu18DORN,Godard17MonoDepth,Ummenhofer17DeMoN}. We also perform the so-called ``cross-dataset'' evaluation task, which tests models trained on a different dataset without fine-tuning. We believe such cross-dataset tasks are essential to
evaluate the robustness and generalization ability~\cite{RobustVisionChallenge18}. Experimental results show that, with reasonably good camera pose estimation, our method outperforms these prior methods on depth estimation with better accuracy, robustness, and temporal stability. Moreover, as shown in Fig.~\ref{fig:teaser}, the output of the proposed method can be directly fed into RGB-D based 3D scanning methods~\cite{Newcombe11KinectFusion, Niessner2013Hashing} for 
3D scene reconstruction.

%RELATED
\section{Related Work}
\label{sec:related}

\paragraph{Depth sensing from active sensors} 

Active depth sensors, such as depth cameras~\cite{horaud16TOF, Remondino13TOF} or LiDAR sensors~\cite{Christian2013ASO} provide dense metric depth measurements as well as sensor-specific confidence measure~\cite{Reynolds-tof-conf11}. 
Despite of their wide usage~\cite{guptaECCV14, Newcombe11KinectFusion, qi2017frustum, Whelan15rss}, they have several inherent drawbacks\cite{chan08, Maimone12, Pomerleau12, Tuley05LIDAR}, such as limited operating range, low spatial resolution, sensor interference, and high power consumption.
Our goal in this paper is to mimic a RGB-D sensor with a monocular RGB camera, which continuously predicts depth (and its uncertainty) from a video stream.

%---------------------------------------------------
\vspace{-1em}
\paragraph{Depth estimation from images}

Depth estimation directly from images has been a core problem in computer vision~\cite{Saxena:2007, Seitz:2006}. 
Classical single view methods~\cite{Criminisi2000, Saxena:2008DRS6} often make strong assumptions on scene structures. 
Stereo and multi-view methods~\cite{Seitz:2006} rely on triangulation and suffer from finding correspondences for textureless regions, transparent/specular materials, and occlusion. 
Moreover, due to global bundle adjustment, these methods are often computationally expensive for real-time applications. 
For depth estimation from a monocular video, there is also scale ambiguity and drifting~\cite{Artal17ORB}. 
Because of these challenges, many computer vision systems~\cite{Artal17ORB, schoenberger2016sfm} use RGB images mainly for camera pose estimation but rarely for dense 3D reconstruction~\cite{schoenberger2016mvs}.
Nevertheless, depth sensing from images has great potentials, since it addresses all the above drawbacks of active depth sensors. In this paper, we take a step in this direction using a learning-based method.

%---------------------------------------------------
\vspace{-1em}
\paragraph{Learning-based depth estimation} 

Recently researchers have shown encouraging results for depth sensing directly from images(s), including single-view methods~\cite{Fu18DORN,Godard17MonoDepth,Zhou17SfmLearner}, video-based methods~\cite{mahjourian2018googleicp,Wang18DDVO,yin2018geonet}, depth and motion from two views~\cite{Chang18PSM,Ummenhofer17DeMoN}, and multi-view stereo~\cite{Huang18DeepMVS,Yao18MVSNet,Zhou18DeepTAM}. 
A few work also incorporated these DL-based depth sensing methods into visual SLAM systems~\cite{Bloesch18CodeSLAM,Tateno17CNNSLAM}.
Despite of the promising performance, however, these DL-based methods are still far from real-world applications, since their robustness and generalization ability is yet to be thoroughly tested~\cite{RobustVisionChallenge18}.
In fact, as shown in Sec.~\ref{sec:results}, we found many state-of-the-art methods degrade significantly even for simple cross-dataset tasks.   
This gives rise to an increasing demand for a systematic study of uncertainty and Bayesian deep learning for depth sensing, as performed in our paper. 
 
%---------------------------------------------------
\vspace{-1em}
\paragraph{Uncertainty and Bayesian deep learning} 

Uncertainty and Bayesian modeling have been long studied in last few decades, with various definitions ranging from the variance of posterior distributions for low-level vision~\cite{Szeliski1990} and motion analysis~\cite{Kim11gprf} to variability of sensor input models~\cite{Kamberova98sensorerrors}. Recently, uncertainty~\cite{Gal2016Dropout,Kendall2017uncertainty} for Bayesian deep learning were introduced for a variety of computer vision tasks~\cite{Clark17VidLoc, Ilg18Uncertainty, Kendall2018multi}. In our work, 
the uncertainty is defined as the posterior probability of depth, \ie, the DPV estimated from a local window of several consecutive frames. Thus, our network estimates the ``measurement uncertainty''~\cite{Kendall2017uncertainty} rather than the ``model uncertainty''. We also learn an additional network module to integrate this depth probability distribution over time in a Bayesian filtering manner, in order to improve the accuracy and robustness for depth estimation from a video stream.

 %--- APPROACH
\vspace{-.5em}
\section{Our Approach}
\label{sec:method}

  \begin{figure*}
     \centering
     \includegraphics[width=0.99\textwidth]{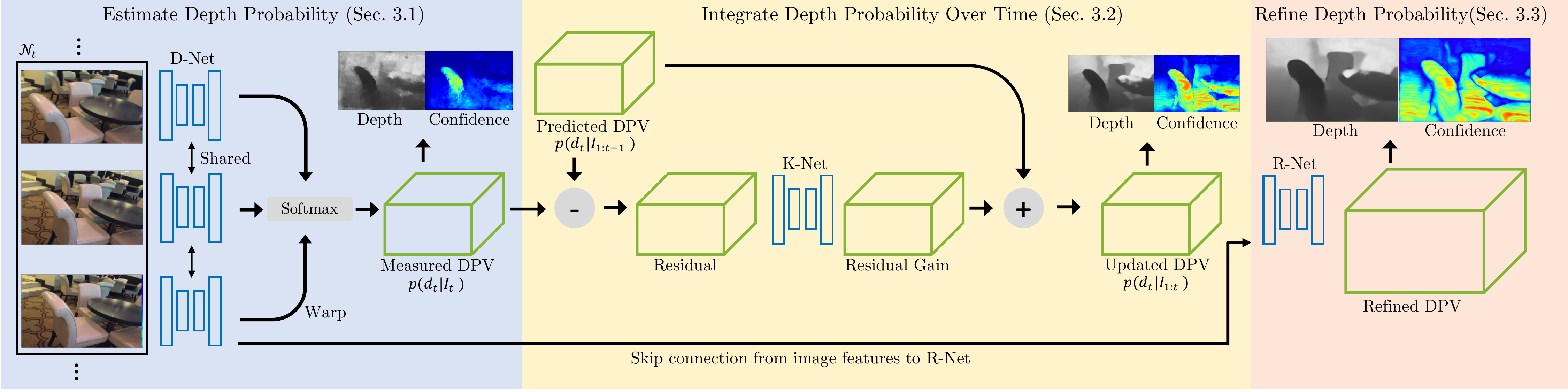}
     \caption{
     Overview of the proposed network for depth estimation with uncertainty from a video. 
     Our method takes the frames in a local time window in the video as input and outputs a Depth Probability Volume (DPV) that is updated over time. 
     The update procedure is in a Bayesian filter fashion: we first take the difference between the local DPV estimated using the D-Net (Sec.~\ref{subsec:method_dnet}) and
     the predicted DPV from previous frames to get the residual; 
     then the residual is modified by the K-Net (Sec.~\ref{subsec:method_kvnet}) and added back to the predicted DPV; 
     at last the DPV is refined and upsampled by the R-Net (Sec.~\ref{subsec:method_refinenet}), which can be used to compute the depth map and its confidence measure.
     }
     \label{fig:network}
 \end{figure*}

Figure~\ref{fig:network} shows an overview of our proposed method for depth sensing from an input video stream, which consists of three parts. The first part (Sec.~\ref{subsec:method_dnet}) is the D-Net, 
which estimates the Depth Probability Volume (DPV) for each input frame. The second part (Sec.~\ref{subsec:method_kvnet}) is the K-Net, 
which helps to integrate the DPVs over time. The third part (Sec.~\ref{subsec:method_refinenet}) is the refinement R-Net, 
which improves the spatial resolution of DPVs with the guidance from input images.  

\begin{figure}
     \centering
     \includegraphics[width=.99\linewidth]{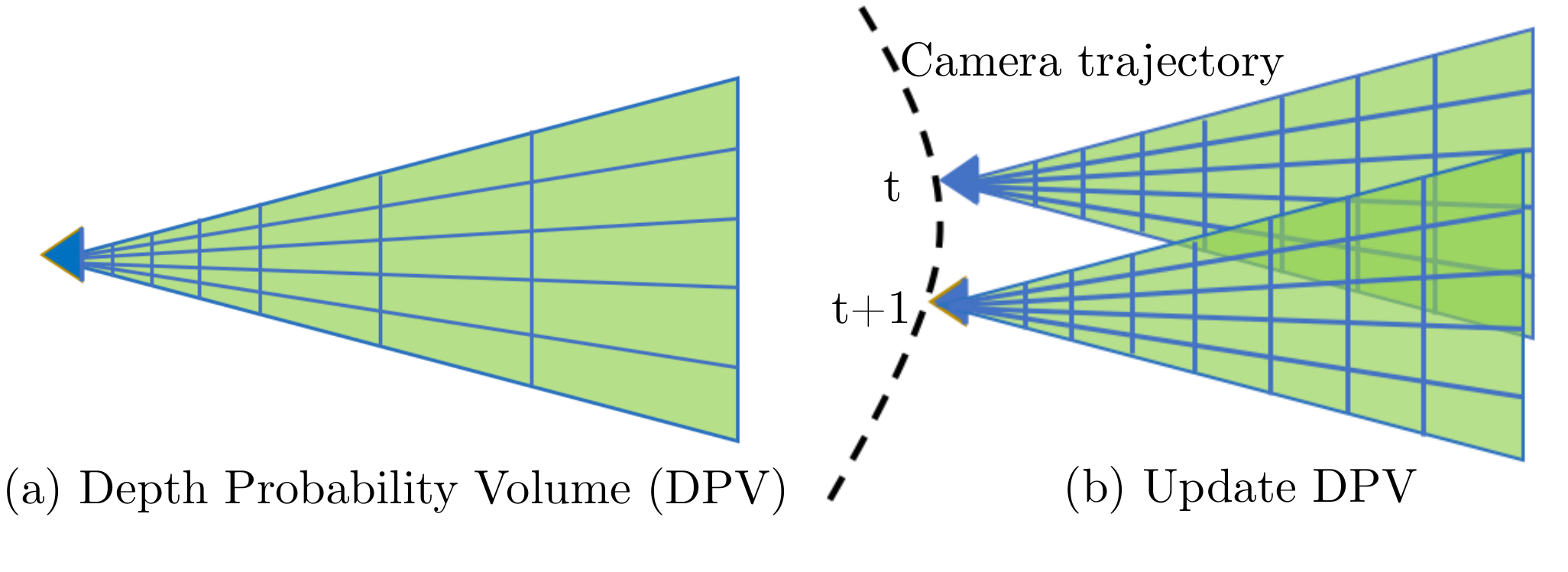}
     \caption{Representation and update for DPV. (a)~The DPV is defined over a 3D frustrum defined by the pinhole camera model . 
     (b)~The DPV gets updated over time as the camera moves. }
     \label{fig:repr}
 \end{figure}

Specifically, we denote the depth probability volume (DPV) as $p(d;u,v)$, which represents the probability of pixel $(u,v)$ having a depth value $d$, where $d\in [d_{min}, d_{max}]$. Due to perspective projection, the DPV is defined on the 3D view frustum attached to the camera, as shown in Fig.~\ref{fig:repr}(a). $d_{min}$ and $d_{max}$ are the near and far planes of the 3D frustum, which is discretized into $N=64$ planes uniformly over the inverse of depth (\ie, disparity). The DPV contains the complete statistical distribution of depth for a given scene. In this paper, we directly use the non-parametric volume to represent DPV. Parametric models, such as Gaussian Mixture Model~\cite{mdn94}, can be also be used.  Given the DPV, we can compute the Maximum-Likelihood Estimates (MLE) for depth and its confidence: 
\begin{align}
    \mbox{Depth}: &\hspace{1em} \hat{d}(u,v) = \sum_{d=d_{min}}^{d=d_{max}} p(d; (u,v))\cdot d, \\
    \mbox{Confidence}: &\hspace{1em} \hat{C}(u,v) = p(\hat{d}, (u,v)).
    \label{eq:dpv_depth}
\end{align} 
To make notations more concise, we will omit $(u,v)$ and use $p(d)$ for DPVs in the rest of the paper.

When processing a video stream, the DPV can be treated as a hidden state of the system. As the camera moves, as shown in Fig.~\ref{fig:repr}(b), the DPV $p(d)$ is being \emph{updated} as new observations arrive, especially for the overlapping volumes. Meanwhile, if camera motion is known, we can easily \emph{predict} the next state $p(d)$ from the current state. This predict-update iteration naturally implies a Bayesian filtering scheme to update the DPV over time for better accuracy. 

%----------------------------------------
\subsection{D-Net: Estimating DPV}
\label{subsec:method_dnet}
 
For each frame $I_t$, we use a CNN, named D-Net, to estimate the conditional DPV, $p(d_t|I_t)$, using $I_t$ and its temporally neighboring frames. 
In this paper, we consider a local time window of five frames $\mathcal{N}_t = [t-2\Delta t,  t-\Delta t, t, t+\Delta t, t+2\Delta t]$, and we set $\Delta t=5$ for all our testing videos (25fps/30fps). For a given depth candidate $d$, we can compute a cost map by warping all the neighboring frames into the current frame $I_t$ and computing their differences. Thus, for all depth candidates, we can compute a cost volume, which produces the DPV after a softmax layer:
\begin{align}
    L(d_t|I_t) &= \sum_{k \in \mathcal{N}_t, k\neq t} ||f(I_t)-\mbox{warp}(f(I_k); d_t, \delta T_{kt})||,\nonumber\\
    p(d_t|I_t) &= \mbox{softmax}(L(d_t|I_t)),
\end{align}
where $f(\cdot)$ is a feature extractor, $\delta T_{kt}$ is the relative camera pose from frame $I_k$ to frame $I_t$, $\mbox{warp}(\cdot)$ is an operator that warps the image features from frame $I_k$ to the reference frame $I_t$, which is implemented as 2D grid sampling. In this paper, without loss of generality, we use the feature extractor $f(\cdot)$ from PSM-Net~\cite{Chang18PSM}, which outputs a feature map  of 1/4 size of the input image. Later in Sec.~\ref{subsec:method_refinenet}, we learn a refinement R-Net to upsample the DPV back to the original size of the input image.

\begin{figure}
     \centering
     \includegraphics[width=0.99\linewidth]{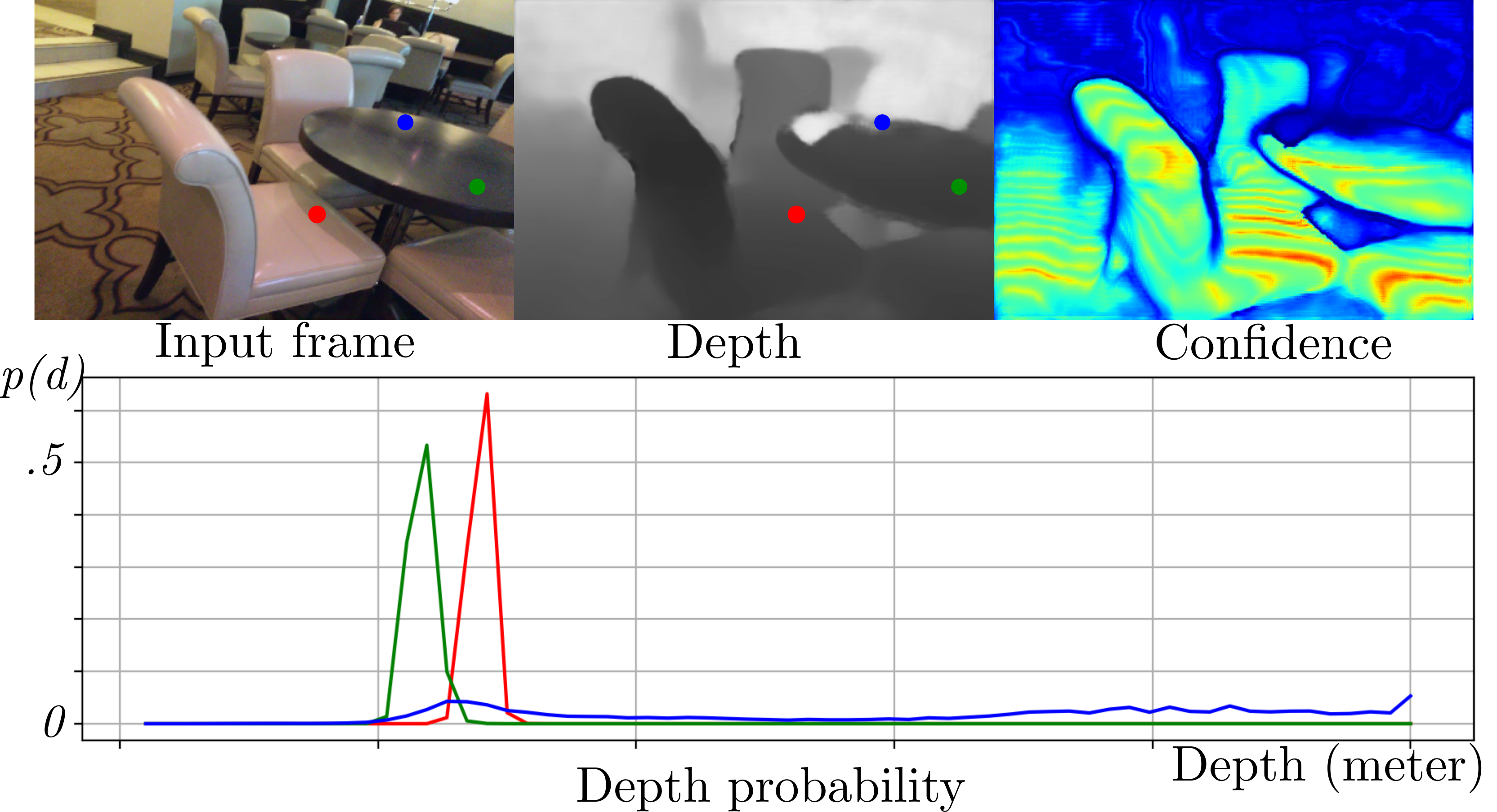}
     \caption{An example of a depth map $\hat{d}(u,v)$ and its confidence map $\hat{C}(u,v)$ (blue means low confidence) derived from a Depth Probability Volume (DPV). The bottom plot shows the depth probability distributions $p(d;u,v)$ for the three selected points, respectively. The red and green points have sharp peaks, which indicates high confidence in their depth values. The blue point is in the highlight region, which results in a flat depth probability distribution and a low confidence for its depth value.} 
     \label{fig:confmap}
 \end{figure}

Figure~\ref{fig:confmap} shows an example of a depth map $\hat{d}(u,v)$ and its confidence map $\hat{C}(u,v)$ (blue means low confidence) derived from a Depth Probability Volume (DPV) from the input image. The bottom plot shows the depth probability distributions $p(d;u,v)$ for the three selected points, respectively. The red and green points have sharp peaks, which indicates high confidence in their depth values. The blue point is in the highlight region, and thus it has a flat depth probability distribution and a low confidence for its depth.

%----------------------------------------
\subsection{K-Net: Integrating DPV over Time}
\label{subsec:method_kvnet}

\begin{figure}
    \centering
    \includegraphics[width=.99\linewidth]{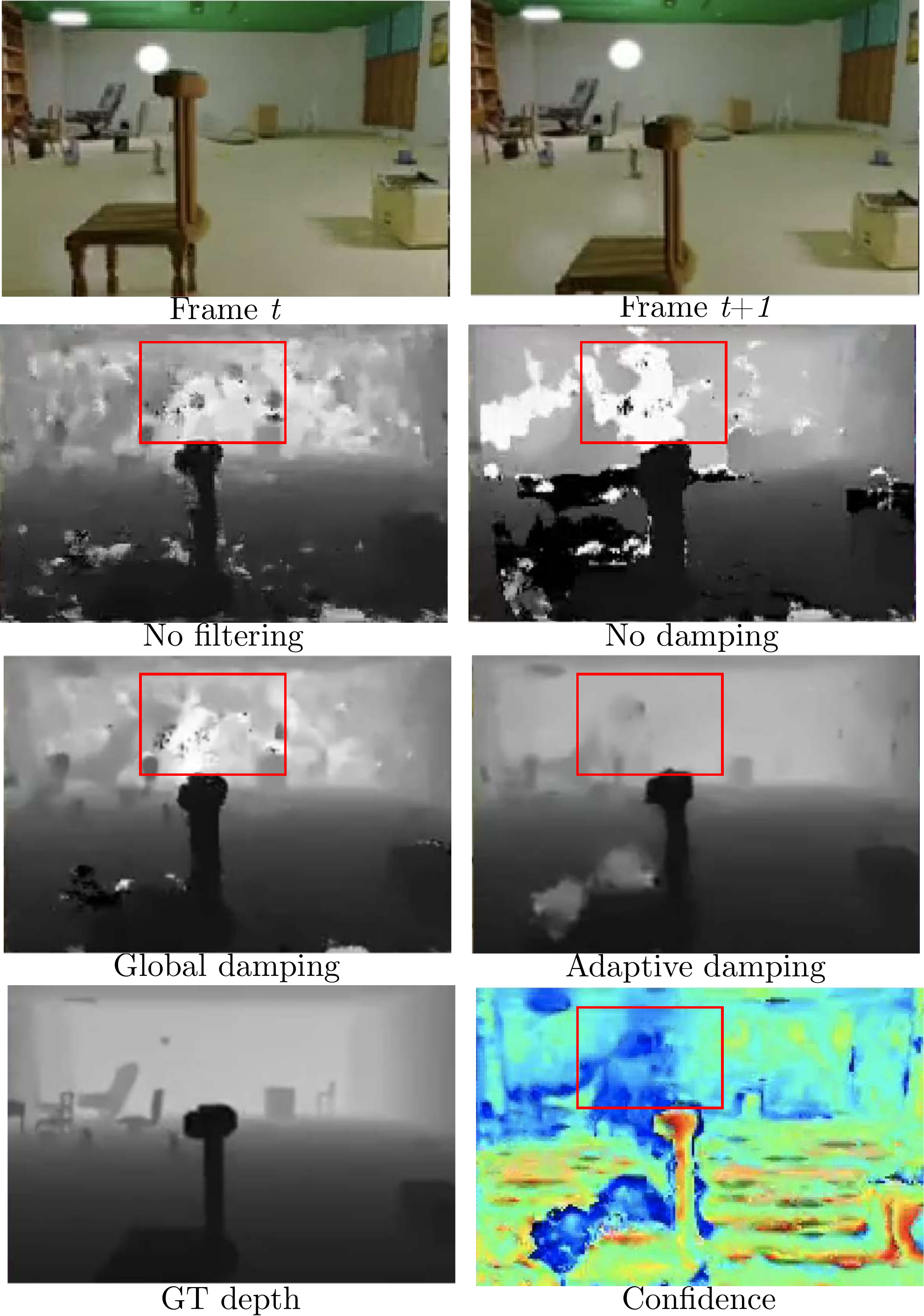}
    \caption{Comparison between different methods for integrating DPV over time. Part of the wall is occluded by the chair at frame $t$ and disoccluded in frame $t+1$.
    \textbf{No filtering}: not integrating the DPV over time.  
    \textbf{No damping}: integrating DPV directly with Bayesian filtering.
    \textbf{Global damping}: down-weighting the predicted DPV for all voxels using Eq.~\ref{eq:global_damping} with $\lambda=0.8$.
    \textbf{Adaptive damping}: down-weighting the predicted DPV adaptively with the K-Net (Sec.~\ref{subsec:method_kvnet}).  
    Using the K-net, we get the best depth estimation for regions with/without disocclusion.
    } 
    \label{fig:damping}
    \vspace{-1em}
\end{figure} 

When processing a video stream, our goal is to integrate the local estimation of DPVs over time to reduce uncertainty. As mentioned earlier, this integration can be naturally implemented as Bayesian filtering. Let us define $d_t$ as the hidden state, which is the depth (in camera coordinates) at frame $I_t$. The ``belief'' volume $p(d_t|I_{1:t})$ is the conditional distribution of the state giving all the previous frames. A simple Bayesian filtering can be implemented in two iterative steps: 
\begin{align}
    \mbox{Predict}:& \hspace{1em} p(d_t | I_{1:t}) \rightarrow p(d_{t+1} | I_{1:t}), \nonumber \\
    \mbox{Update}:& \hspace{1em}  p(d_{t+1} | I_{1:t}) \rightarrow p(d_{t+1} | I_{1:t+1}),
\end{align}
where the prediction step is to warp the current DPV from the camera coordinate at $t$ to the camera coordinate at $t+1$: 
\begin{equation}
     p(d_{t+1}| I_{1:t}) = \mbox{warp}(p(d_t|I_{1:t}), \delta T_{t,t+1}),
     \label{eq:bayesian_p}
\end{equation}
where $\delta T_{t,t+1}$ is the relative camera pose from time $t$ to time $t+1$, and $\mbox{warp}(\cdot)$ here is a warping operator implemented as 3D grid sampling. At time $t+1$, we can compute the local DPV $p(d_{t+1}|I_{t+1})$ from the new measurement $I_{t+1}$ using the D-Net. This local estimate is thus used to update the hidden state, \ie, the ``belief'' volume, 
\begin{equation}
    \label{eq:bayesian}
    p(d_{t+1} | I_{1:t+1}) = p(d_{t+1} | I_{1:t}) \cdot p(d_{t+1}|I_{t+1}).
\end{equation}
Note that we always normalize the DPV in the above equations and ensure $\int_{d_{min}}^{d_{max}} p(d)=1$. Figure~\ref{fig:damping} shows an example. As shown in the second row, with the above Bayesian filtering (labeled as "no damping"), the estimated depth map is less noisy, especially in the regions of the back wall and the floor. 

However, one problem with directly applying Bayesian filtering is it integrates both correct and incorrect information in the prediction step. For example, when there are occlusions or disocclusions, the depth values near the occlusion boundaries change abruptly. Applying Bayesian filtering directly will propagate wrong information to the next frames for those regions, as highlighted in the red box in Fig.~\ref{fig:damping}. One straightforward solution is to reduce the weight of the prediction in order to prevent incorrect information being integrated over time. Specifically, by defining $E(d)=-\log p(d)$, Eq.~\ref{eq:bayesian} can be re-written as 
\begin{equation}
    E(d_{t+1} | I_{1:t+1} ) = E(d_{t+1} | I_{1:t}) + E(d_{t+1} | I_{t+1}), \nonumber
\end{equation}
where the first term is the prediction and the second term is the measurement. To reduce the weight of the prediction, we multiply a weight $\lambda\in [0,1]$ with the first term,
\begin{equation}
    E(d_{t+1} | I_{1:t+1} ) = \lambda\cdot E(d_{t+1} | I_{1:t}) + E(d_{t+1} | I_{t+1}).
    \label{eq:global_damping}
\end{equation}
We call this scheme ``global damping''. As shown in Fig.~\ref{fig:damping}, global damping helps to reduce the error in the disoccluded regions. However, global damping may also prevent some correct depth information to be integrated to the next frames, since it reduces the weights equally for all voxels in the DPV. Therefore, we propose an ``adaptive damping'' scheme to update the DPV: 
\begin{equation} 
   E(d_{t+1} | I_{1:t+1} ) =  E(d_{t+1} | I_{1:t}) + g(\Delta E_{t+1}, I_{t+1}),  
   \label{eq:kvnet}
\end{equation}
where $\Delta E_{t+1}$ is the difference between the measurement and the prediction,
\begin{equation}
    \Delta E_{t+1} =   E(d_{t+1}| I_{t+1}) - E(d_{t+1} | I_{1:t}),
    \label{eq:kvnet_residual}
\end{equation} 
and $g(\cdot)$ is a CNN, named K-Net, which learns to transform $\Delta E_{t+1}$ into a correction term to the prediction. Intuitively, for regions with correct depth probability estimates, the values in the overlapping volume of DPVs are consistent. Thus the residual in Eq.~\ref{eq:kvnet_residual} is small and the DPV will not be updated in Eq.~\ref{eq:kvnet}. On the other hand, for regions with incorrect depth probability, the residual would be large and the DPV will be corrected by $g(\Delta E, I_{t+1})$. This way, the weight for prediction will be changed adaptively for different DPV voxels. As shown in Fig.~\ref{fig:damping}, the adaptive damping, \ie, K-Net, significantly improve the accuracy for depth estimation. In fact, K-Net is closely related to the derivation of Kalman filter, where ``K'' stands for Kalman gain. Please refer to the appendix for details.

%----------------------------------------
\subsection{R-Net and Training Details}
\label{subsec:method_refinenet}

Finally, since the DPV $p(d_t|I_{1:t})$ is estimated with $1/4$ spatial resolution (on both width and height) of the input image, we employ a CNN, named R-Net, to upsample and refine the DPV back to the original image resolution. The R-Net, $h(\cdot)$, is essentially an U-Net with skip connections, which takes input the low-res DPV from the K-Net $g(\cdot)$ and the image features extracted from the feature extractor $f(\cdot)$, and outputs a high-resolution DPV.

In summary, as shown in Fig.~\ref{fig:network}, the entire network has three modules, \ie, the D-Net, $f(\cdot;\Theta_1)$, the K-Net, $g(\cdot; \Theta_2)$, and the R-Net, $h(\cdot;\Theta_3)$. Detailed network architectures are provided in the appendix.  The full network is trained end-to-end, with simply the Negative Log-Likelihood (NLL) loss over the depth, $\mbox{Loss} = \mbox{NLL}(p(d), d_{GT})$. We also tried to add image warping as an additional loss term (\ie, minimizing the difference between $I_t$ and the warped neighboring frames), but we found that it does not improve the quality of depth prediction. 

During training, we use ground truth camera poses. For all our experiments, we use the ADAM optimizer \cite{Diederik2018Adam} with a learning rate of $10^{-5}$, $\beta_1=.9$ and  $\beta_2=.999$. The whole framework, including D-Net, K-Net and R-Net, is trained together in an end-to-end fashion for 20 epochs.  

%----------------------------------------
\subsection{Camera Poses during Inference}
\label{subsec:method_lba} 

During inference, given an input video stream, our method requires relative camera poses $\delta T$ between consecutive frames --- at least for all the first five frames --- to bootstrap the computation of the DPV. In this paper, we evaluated several options to solve this problem. In many applications, such as autonomous driving and AR, initial camera poses may be provided by additional sensors such as GPS, odometer, or IMU. Alternatively, we can also run state-of-the-art monocular visual odometry methods, such as DSO~\cite{Engel18pami}, to obtain the initial camera poses. Since our method outputs continuous dense depth maps and their uncertainty maps, we can in fact further optimize the initial camera poses within a local time window, similar to local bundle adjustment~\cite{Triggs99}. 
 
\begin{figure}
    \centering
    \includegraphics[width=.99\linewidth]{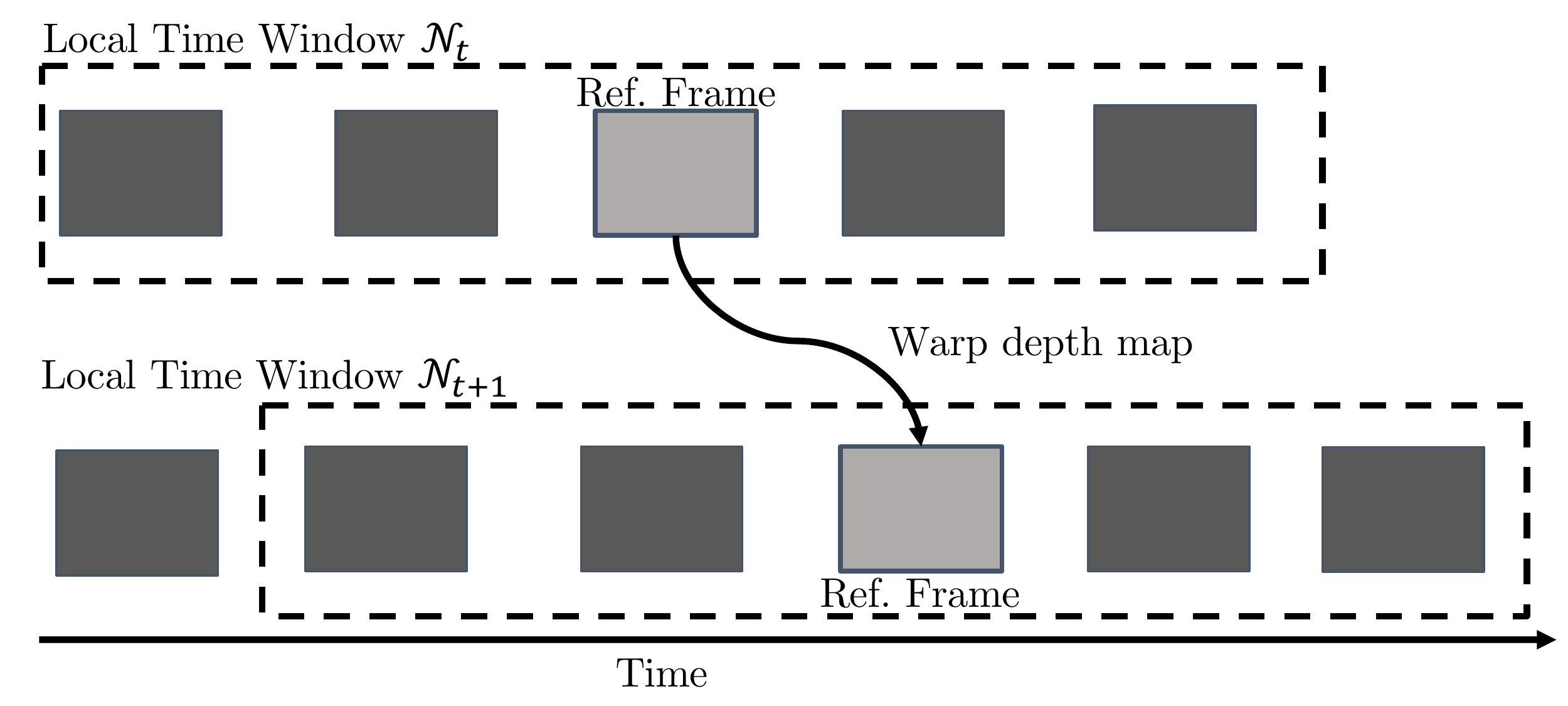}
    \caption{Camera pose optimization in a sliding local time window during inference. Given the relative camera pose from the reference frame in $\mathcal{N}_t$ to the reference frame in $\mathcal{N}_{t+1}$, we can predict the depth map for the reference frame in $\mathcal{N}_{t+1}$. Then, we optimize the relative camera poses between every source frame and the reference frame in $\mathcal{N}_{t+1}$ using Eq.\ref{eq:opt_pose}.}
    \label{fig:pose_opt}
    \vspace{-1em}
\end{figure} 

Specifically, as shown in Fig.~\ref{fig:pose_opt}, given $p(d_{t}|I_{1:t})$, the DPV of the reference frame $I_t$ in the local time window $\mathcal{N}_{t}$, we can warp $p(d_t|I_{1:t})$ to the reference camera view in $\mathcal{N}_{t+1}$ to predict the DPV $p(d_{t+1}|I_{1:t})$ using Eq.~\ref{eq:bayesian_p}.
Then we get the depth map $\hat{d}$ and confidence map $\hat{C}$ for the new reference frame using Eq.~\ref{eq:dpv_depth}. 
The camera poses within the local time window $\mathcal{N}_{t+1}$ are optimized as:
\begin{equation}
\begin{aligned}
  \vspace{1em} 
\underset{ \substack{\delta T_{k,t+1}  \\
k \in \mathcal{N}_{t+1}, k\neq t+1} }{\text{min.}}
 \vspace{1em} \sum_k  \hat{C}
|I_{t+1} - \text{warp}(I_{k}; \hat{d}; \delta T_{k, t+1}) |_1,
\end{aligned}
\label{eq:opt_pose}
\end{equation}
where $\delta T_{k,t+1}$ is the relative camera pose of frame $k$ to frame $t+1$;
$I_{k}$ is the source image at frame $k$;  
$\mbox{warp}(\cdot)$ 
is an operator that warps the image from the source view to the reference view.

%-- RESULTS 
\vspace{-0.5em}
\section{Experimental Results}
\label{sec:results}
 
\begin{figure*}
    \centering
    \includegraphics[width=\linewidth]{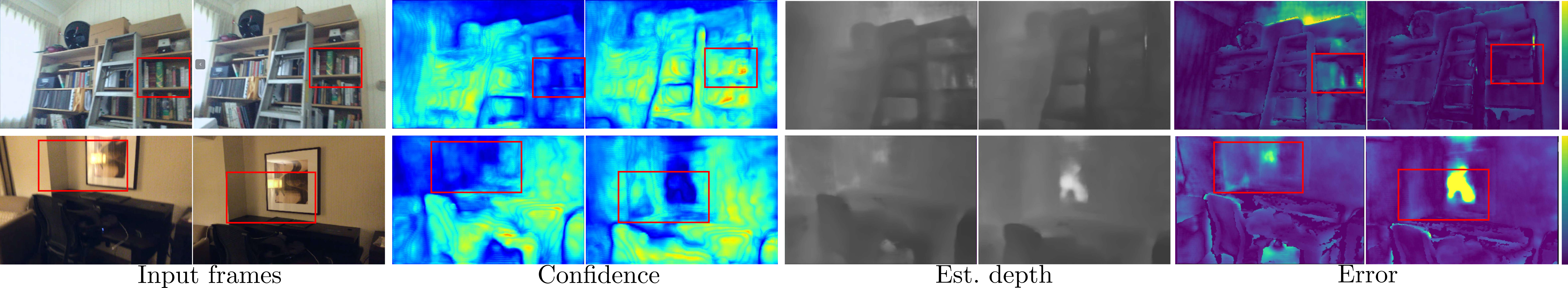}
    \caption{Exemplar results of our approach on ScanNet~\cite{dai2017scannet}. In addition to high quality depth output, we also obtain reasonable confidence maps (as shown in the marked regions for occlusion and specularity) which correlates with the depth error. Moreover, the confidence maps accumulate correctly over time with more input frames.}
    \label{fig:uncertainty}
    \vspace{-0.5em}
\end{figure*}
 
We evaluate our method on multiple indoor and outdoor datasets~\cite{shotton13data,Sturm12iros,Gaidon16cvpr,Geiger12cvpr}, 
with an emphasis on accuracy and robustness. For accuracy evaluation, we argue the widely-used statistical metrics~\cite{Eigen14depth, Ummenhofer17DeMoN} are insufficient because they can only provide an overall estimate over the entire depth map. Rather, we feed the estimated depth maps directly into classical RGB-D based 3D scanning systems~\cite{Newcombe11KinectFusion,Niessner2013Hashing} for 3D reconstruction --- this will show the metric accuracy, the consistency, and the usefulness of the estimation. For robustness evaluation, we performed the aforementioned cross-dataset evaluation tasks, \ie, testing on new datasets without fine-tuning. The performance degradation over new datasets will show the generalization ability and robustness for a given algorithm. 

As no prior work operates in the exact setting as ours, it is difficult to choose methods to compare with. We carefully select a few recent DL-based depth estimation methods and try our best for a fair comparison. For single-view methods, we select DORN~\cite{Fu18DORN} which is the current state-of-the-art~\cite{RobustVisionChallenge18}. For two-view methods, we compare with DeMoN~\cite{Ummenhofer17DeMoN}, which shows high quality depth prediction from a pair of images. We also compare with MonoDepth~\cite{Godard17MonoDepth}, which is a semi-supervised learning approach from stereo images. 
To improve the temporal consistency for these per-frame estimations, we trained a post-processing network~\cite{Lai18Temporal}, but we observed it does not improve the performance.
Since there is always scale ambiguity for depth from a monocular camera, for fair comparison, we normalize the scale for the outputs from all the above methods before we compute statistical metrics~\cite{Eigen14depth}.

The inference time for processing one frame in our method is $\sim$ $0.7$ second per frame without pose optimization and $\sim$ $1.5$ second with pose estimation on a workstation with GTX 1080 GPU and 64 GB RAM memory, with the framework implemented in Python.  The pose estimation part can be implemented with C++ to improve efficiency.

%--------------------------------------------
\vspace{-1em}
\paragraph{Results for Indoor Scenarios} 

\begin{table}[t]
    \centering
    \caption{Comparison of depth estimation over the 7-Scenes dataset~\cite{shotton13data} with the metrics defined in~\cite{Eigen14depth}.} 
    \begin{tabular}{ rcccc }
    \toprule
  &  $\sigma<1.25$  
  & abs. rel  
  & rmse   
  & scale inv. \\ 
\midrule
DeMoN~\cite{Ummenhofer17DeMoN}  
& 31.88 
& 0.3888 
& 0.8549 
& 0.4473\\  
DORN~\cite{Fu18DORN}  
& 60.05 
& 0.2000 
& 0.4591 
& 0.2207 \\  
Ours 
& \textbf{69.26} 
& \textbf{0.1758} 
& \textbf{0.4408} 
& \textbf{0.1899}\\  
\bottomrule
    \end{tabular}
    \label{tab:result_7scenes}
    \vspace{-1em}
\end{table}

We first evaluated our method for indoor scenarios, for which RGB-D sensors were used to capture dense metric depth for ground truth. We trained our network on ScanNet~\cite{dai2017scannet}. 
Figure~\ref{fig:uncertainty} shows two exemplar results. As shown, in addition to depth maps, our method also outputs reasonable confidence maps (\eg, low confidence in the occluded or specular regions) which correlates with the depth errors. 
Moreover, with more input frames, the confidence maps accumulate correctly over time: the confidence of the books (top row) increases and the depth error decreases; 
the confidence of the glass region (bottom row) decreases and the depth error increases.   
    
For comparison, since the models provided by DORN and DeMoN were trained on different datasets, we compare with these two methods on a separate indoor dataset 7Scenes~\cite{shotton13data}. For our method, we assume that the relative camera rotation $\delta R$ within a local time window is provided (\textit{e.g.} measured by IMU). As shown in Table~\ref{tab:result_7scenes}, our method significantly outperforms both DeMoN and DORN on this dataset based on the commonly used statistical metrics~\cite{Eigen14depth}.  We include the complete metrics in the appendix.
Without using an IMU, our method can also achieve better performance, as shown in Table~\ref{tab:ablation_pose}.

\begin{figure*}
    \centering
    \includegraphics[width=.99\linewidth]{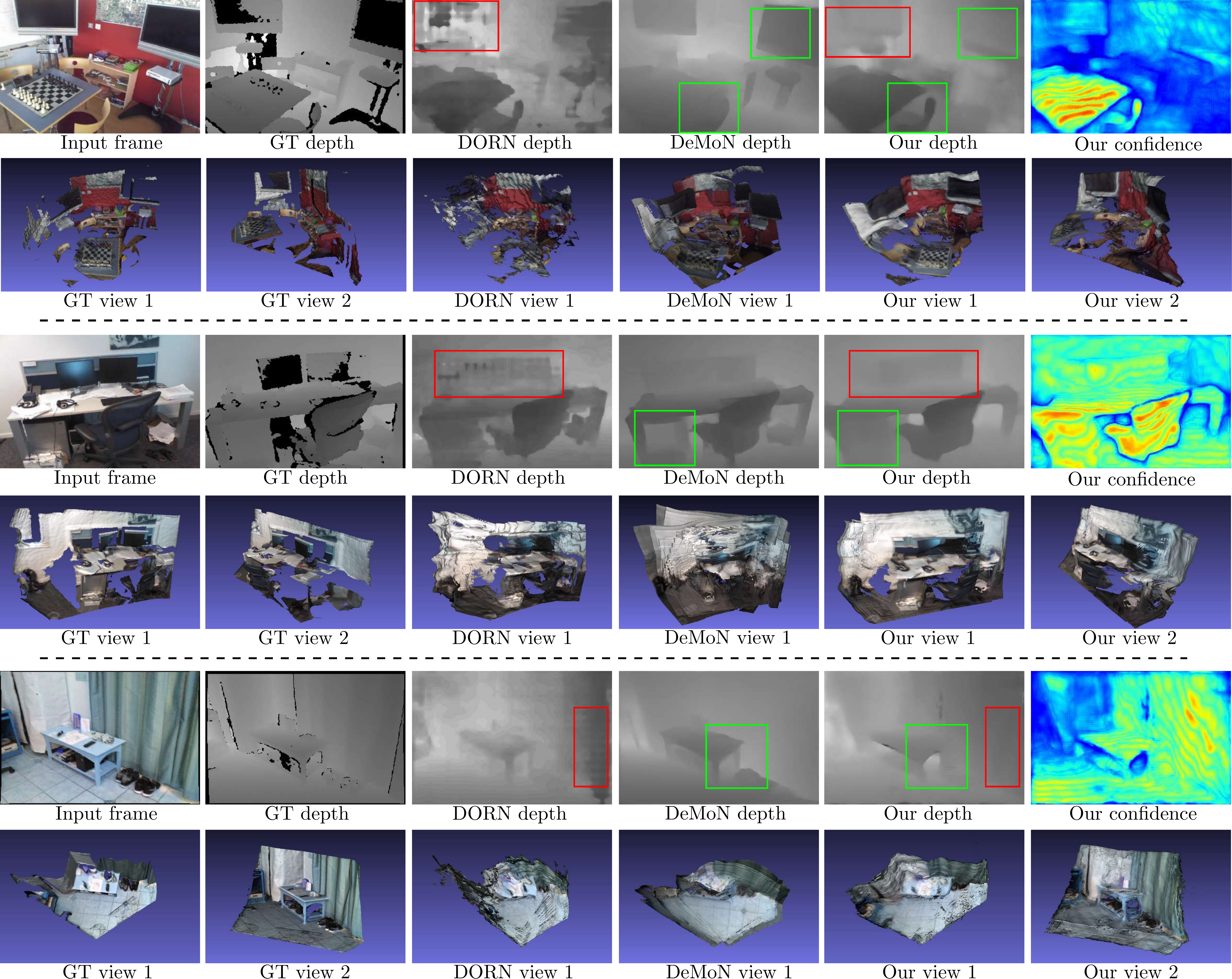}
    \caption{Depth and 3D reconstruction results on indoor datasets (best viewed when zoomed in).
    We compare our method with DORN \cite{Fu18DORN} and DeMoN \cite{Ummenhofer17DeMoN}, in terms of both depth maps and 3D reconstruction using Voxel Hashing \cite{Niessner2013Hashing} that accumulates the estimated depth maps for multiple frames.
    To show the temporal consistency of the depths, we use different numbers of depth maps for Voxel Hashing: $2$ depth maps for the first sample and $30$ depth maps for the other samples.
    The depth maps from DORN contain block artifacts as marked in red boxes. This is manifested as the rippled shapes in the 3D reconstruction.
    DeMoN generates sharp depth boundaries but fails to recover the depth faithfully in the regions marked in the green box. Also, the depths from DeMoN is not temporally consistent. This leads to the severe misalignment artifacts in the 3D reconstructions.
    In comparison, our method generates correct and temporally consistent depths maps, especially at regions with high confidence, such as the monitor where even the Kinect sensor fails to get the depth due to low reflectance.}
    \label{fig:result_3d}
    \vspace{-1em}
\end{figure*}

For qualitative comparison, as shown in Fig.~\ref{fig:result_3d}, the depth maps from our method are less noisy, more sharper, and temporally more consistent. More importantly, using an RGB-D 3D scanning method~\cite{Niessner2013Hashing}, we can reconstruct a much higher quality 3D mesh with our estimated depths compared to other methods. Even when compared with 3D reconstruction using a real RGB-D sensor, our result has better coverage and accuracy in some regions (\eg, monitors / glossy surfaces) where active depth sensors cannot capture.

%--------------------------------------------
\vspace{-1em}
\paragraph{Results for Outdoor Scenarios}

We also evaluated our method on some outdoor datasets --- KITTI~\cite{Geiger12cvpr} and virtual KITTI~\cite{Gaidon16cvpr}. The virtual KITTI dataset is used because it has dense, accurate metric depth as ground truth, while KITTI only has sparse depth values from LiDAR as ground truth. For our method, we use the camera poses measured by the IMU and GPS.
Table~\ref{tab:result_kitti} lists the comparison results with DORN~\cite{Fu18DORN}, Eigen~\cite{Eigen14depth}, and MonoDepth~\cite{Godard17MonoDepth} which are also trained on KITTI~\cite{Geiger12cvpr}. Our method has similar performance with DORN~\cite{Fu18DORN}, and is better than the other two methods, based on the statistical metrics defined in~\cite{Eigen14depth}. We also tested our method with camera poses from DSO~\cite{Engel18pami} and obtain slightly worse performance (see appendix). 

Figure~\ref{fig:result_kitti} shows qualitative comparison for depth maps in KITTI dataset. As shown, our method generate sharper and less noisier depth maps. In addition, our method outputs depth confidence maps (\eg, lower confidence on the car window). Our depth estimation is temporally consistent, which leads to the possibility of fusing multiple depth maps with voxel hashing~\cite{Niessner2013Hashing} in the outdoors for a large-scale dense 3D reconstruction, as shown in Fig.~\ref{fig:result_kitti}.
 
In Table~\ref{tab:result_vkitti}, we performed the cross-dataset task. The left shows the results with training from KITTI~\cite{Geiger12cvpr} and testing on virtual KITTI~\cite{Gaidon16cvpr}. The right shows the results with training from indoor datasets (NYUv2~\cite{Silberman12ECCV} for DORN~\cite{Fu18DORN} and ScanNet~\cite{dai2017scannet} for ours) and testing on KITTI~\cite{Geiger12cvpr}. As shown, our method performs better, which shows its better robustness and generalization ability.

\begin{table}
    \centering
    \caption{Comparison of depth estimation on KITTI~\cite{Geiger12cvpr}.} 
    \begin{tabular}{rcccc} 
    \toprule
  &  $\sigma<1.25$  
  & abs. rel  
  & rmse   
  & scale inv. \\ 
 \midrule
Eigen~\cite{Eigen14depth} 
& 67.80 
& 0.1904 
& 5.114 
& 0.2628
\\  
Mono~\cite{Godard17MonoDepth}  
& 86.43  
& 0.1238 
& 2.8684 
& 0.1635
\\  
DORN~\cite{Fu18DORN}  
& 92.62  
& \textbf{0.0874}  
& 3.1375 
& 0.1233  \\  
Ours 
& \textbf{93.15}  
& 0.0998  
& \textbf{2.8294} 
& \textbf{0.1070} \\ 
\bottomrule
    \end{tabular}
    \label{tab:result_kitti}
    \vspace{-1em}
\end{table}

\begin{figure*}
    \centering
    \includegraphics[width=\linewidth]{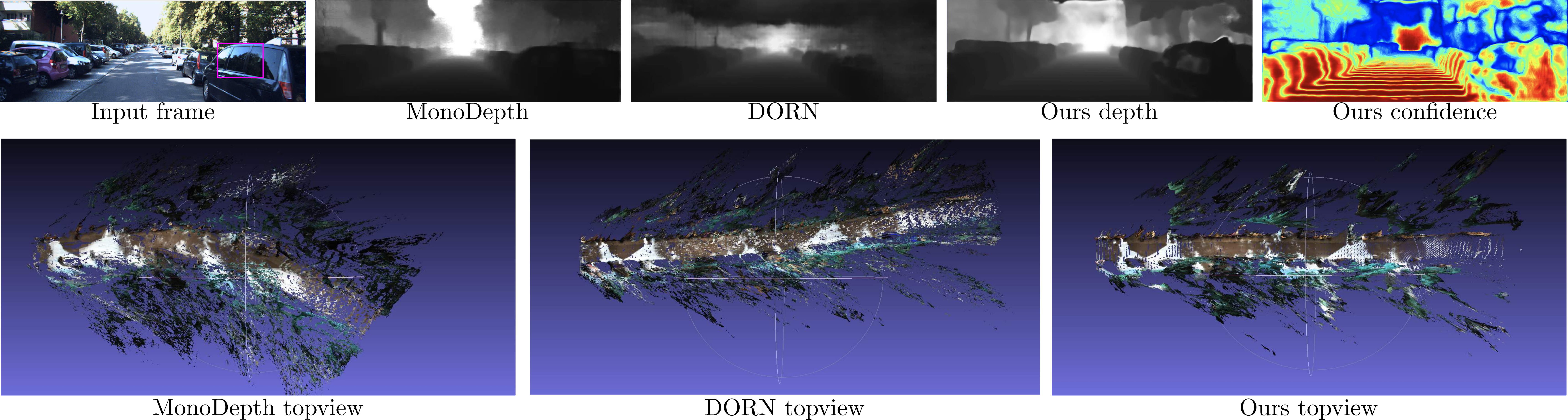}
    \caption{ 
    Depth map and 3D reconstruction for KITTI, compared with DORN \cite{Fu18DORN}, MonoDepth \cite{Ummenhofer17DeMoN} (best viewed when zoomed in).
    First row: Our depth map is sharper and contains less noise.   
    For specular region (marked in the pink box), the confidence is lower. 
    Second row, from left to right: reconstructions using depth maps of the same 100 frames estimated from MonoDepth, DORN and our method. All meshes are viewed from above. Within the 100 frames, the vehicle was travelling in a straight line without turning.  
    }
    \label{fig:result_kitti}
\end{figure*}

\begin{table}
    \vspace{-1em}
    \centering
    \caption{Cross-dataset tests for depth estimation in the outdoors.}
    \begin{tabular}{rcccc} 
    \toprule
        & \multicolumn{4}{c}{KITTI (train) $\rightarrow$ virtual KITTI (test)}\\
    \cmidrule(r){2-5} 
        &  $\sigma<1.25$  &  abs. rel  &  rmse  &  scale inv.\\
    \midrule  
    DORN~\cite{Fu18DORN}  & 69.61 & \textbf{0.2256}  & 9.618 & 0.3986 \\  
    Ours  & \textbf{73.38} & 0.2537  & \textbf{6.452} & \textbf{0.2548} \\
    \midrule
      & \multicolumn{4}{c}{Indoor (train) $\rightarrow$ KITTI (test)} \\
   \cmidrule(r){2-5} 
        &  $\sigma<1.25$  &  abs. rel  &  rmse  &  scale inv.\\
   \midrule  
    DORN~\cite{Fu18DORN} & 25.44 & 0.6352 & 8.603 & 0.4448\\
    Ours & \textbf{72.96} & \textbf{0.2798} & \textbf{5.437}  & \textbf{0.2139}\\  
    \bottomrule
    \end{tabular}
    \label{tab:result_vkitti}
    \vspace{-.5em}
\end{table}

%--------------------------------------------
\vspace{-1em}
\paragraph{Ablation Study}

The performance of our method relies on accurate estimate of camera poses, so we test our method with different camera pose estimation schemes:
(1) Relative camera rotation $\delta R$ is read from an IMU sensor (denoted as ``GT $R$''). 
(2) $\delta R$ of all frames are initialized with DSO~\cite{Engel18pami} (denoted as ``VO pose'')
(3) $\delta R$ of the first five frames are initialized with DSO~\cite{Engel18pami} (denoted as ``$1$\textit{st} win'').  
We observe that when only the camera poses in the first time window are initialized using DSO, the performance in terms of depth estimation is better than that using the DSO pose initialization for all frames. This may seem counter-intuitive, but it is because monocular VO methods sometimes have large errors for textureless regions while optimization with dense depths may overcome this problem.

\begin{table}
    \centering
    \caption{Performance on 7Scenes with different initial poses}
    \begin{tabular}{rcccc} 
    \toprule
  &  $\sigma<1.25$  
  & abs. rel  
  & rmse   
  & scale inv.  \\ 
 \midrule
VO pose 
& 60.63
& 0.1999
& 0.4816
& 0.2158
\\  
$1$\textit{st} win.
& 62.08
& 0.1923
& 0.4591
& 0.2001
\\
GT $R$
& 69.26
& 0.1758
& 0.4408
& 0.1899
\\ 
GT pose 
& 70.54
& 0.1619
& 0.3932
& 0.1586  \\
\bottomrule
    \end{tabular}
    \label{tab:ablation_pose}
    \vspace{-1em}
\end{table}   

%--------------------------------------------
\vspace{-1em}
\paragraph{Usefulness of the Confidence Map}

\begin{figure}
    \centering
    \includegraphics[width=.95\linewidth]{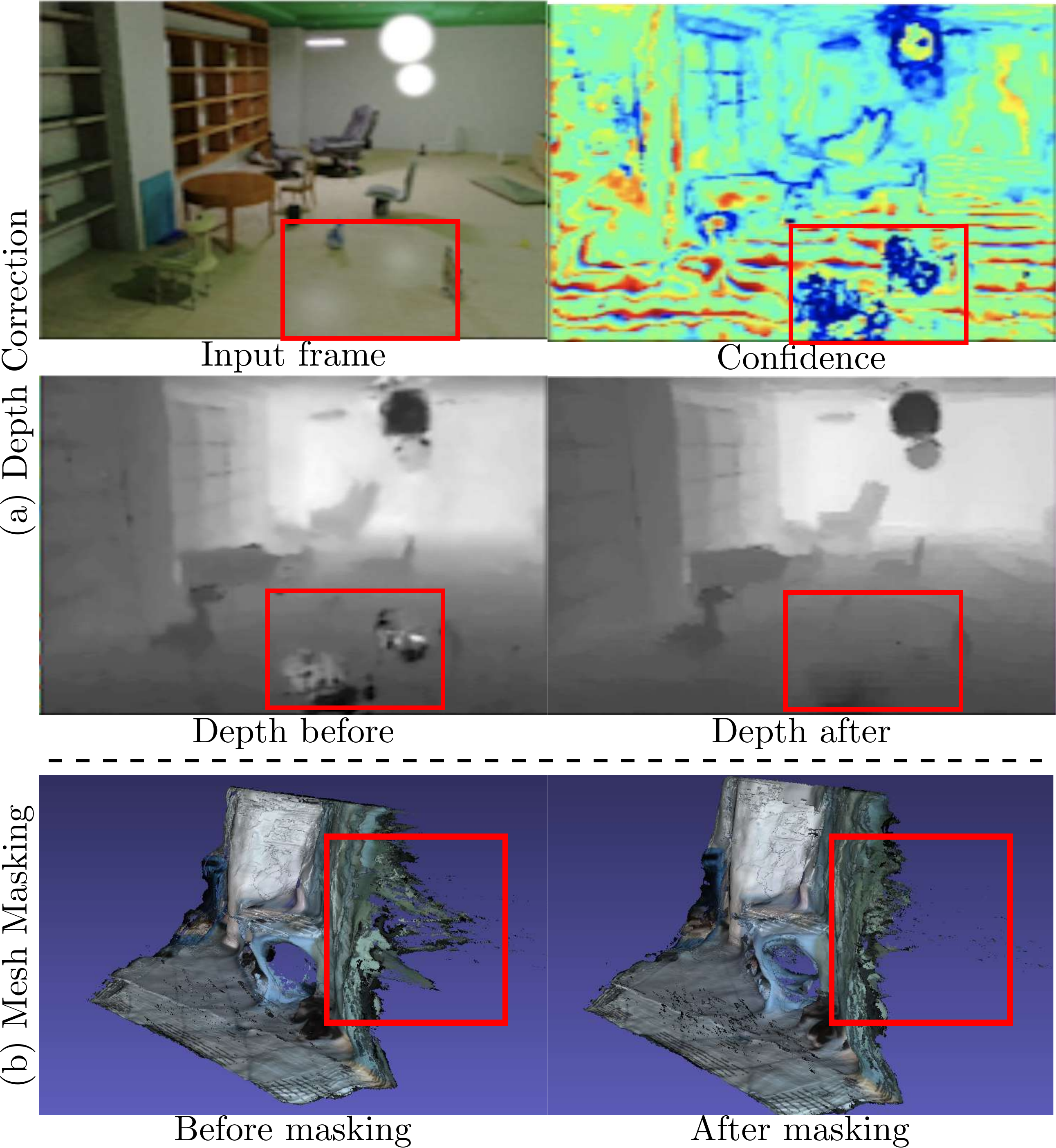}
    \caption{Usefulness of depth confidence map. (a) Correct depth map using Fast Bilateral Solver \cite{BarronPoole2016}. 
    (b) Mask out pixels with low confidence before applying Voxel Hashing \cite{Niessner2013Hashing}. 
    }
    \label{fig:conf_FBS}
    \vspace{-2em}
\end{figure}

The estimated confidence maps can also be used to further improve the depth maps. As shown in Fig.~\ref{fig:conf_FBS}(a), given the depth map and the corresponding confidence, we can correct the regions with lower confidence due to specular reflection. Also, for 3D reconstruction algorithm, given the depth confidence, we can mask out the regions with lower confidence for better reconstruction, as shown in Fig.~\ref{fig:conf_FBS}(b). 

%-- CONCLUSION 
\vspace{-.5em}
\section{Conclusions and Limitations}
\label{sec:conclusion}

In this paper, we present a DL-based method for continuous depth sensing from a monocular video camera. Our method estimates a depth probability distribution volume from a local time window, and integrates it over time under a Bayesian filtering framework. 
Experimental results show our approach achieves high accuracy, temporal consistency, and robustness for depth sensing, especially for the cross-dataset tasks. The estimated depth maps from our method can be fed directly into RGB-D scanning systems for 3D reconstruction and achieve on-par or sometimes more complete 3D meshes than using a real RGB-D sensor.

There are several limitations that we plan to address in the future. First, camera poses from a monocular video often suffer from scale drifting, which may affect the accuracy of our depth estimation. 
Second, in this work we focus on depth sensing from a local time window, rather than solving it in a global context using all the frames.  

\newpage

\begin{appendices}

\section{Relation of K-Net to the Kalman filter}
The proposed update process defined in Eq.~8 in the main paper using residuals is closely related to Kalman Filter.
In Kalman Filter, given the observation $x_t$ at time $t$ and the estimated hidden state $h_{t-1}$ at time $t-1$, the updated hidden state $h_{t}$ is: 
\begin{equation} 
    h_t = W_t h_{t-1}   + K_t(x_t - V_t W_t h_{t-1})  
    \label{eq:kvnet_kalman}
\end{equation}
where $W_t$ is the transition matrix mapping the previous hidden state to current state; 
$K_t$ is the gain matrix mapping the residual in the observation space to the hidden state space. 
$V_t$ is the measurement matrix mapping the estimation in the hidden state space back to the observation space. 

If we assume the measurement matrix is accurate: $x_t = V h_t$, and the gain and measurement matrices are temporally invariant, 
we have:
\begin{align}
     h_t   &= W_t h_{t-1} 
         + K(V h_t - V W_t h_{t-1}) \nonumber \\
          &= W_t h_{t-1} 
         + K V( h_t -  W_t h_{t-1})
    \label{eq:kvnet_kalman2}
\end{align} 
Comparing our proposed update process in Eq.~5, Eq.~8 and Eq.~9 in the main paper
and
Kalman Filter in Eq.\ref{eq:kvnet_kalman2}, 
in our case the input images correspond to the observations $x_t$ ;
the negative-log depth probabilities correspond to the hidden states $h_t$;
the warping operator $\text{warp}(\cdot)$ corresponds to the transition matrix $W_t$; 
the K-Net $g(\cdot)$ corresponds to the multiplication of the gain and measurement matrices $KV$
in Eq.~\ref{eq:kvnet_kalman2}. 

\section{More Results}
\subsection{Complete metrics for Comparisons} 
We show the
complete metrics for depth estimation comparisons
in Table~\ref{tab:result_7scenes}
and Table~\ref{tab:result_kitti}. 
\begin{table*}[!t]
    \centering
    \caption{Comparison of depth estimation over the 7-Scenes dataset~\cite{shotton13data} with the metrics defined in~\cite{Eigen14depth}} 
    \begin{tabular}{ m{4em}  m{4em}   m{4.5em}   m{4.5em}  m{4em}   m{4em}   m{4em}  m{4em}   m{4em}  }
    \toprule
  &  $\sigma<1.25$ 
  &  $\sigma<1.25^2$ 
  &  $\sigma<1.25^3$
  & abs. rel 
  & sq. rel
  & rmse  
  & rmse log
  & scale. inv \\ 
\midrule
DeMoN~\cite{Ummenhofer17DeMoN}  
& 31.88
& 61.02
& 82.52
& 0.3888
& 0.4198
& 0.8549
& 0.4771
& 0.4473\\  
DORN~\cite{Fu18DORN}  
& 60.05
& 87.76
& 96.33
& 0.2000
& 0.1153
& 0.4591
& 0.2813
& 0.2207 \\  
Ours  
& \textbf{69.26}
& \textbf{91.77}
& \textbf{96.82}
& \textbf{0.1758}
& \textbf{0.1123}
& \textbf{0.4408}
& \textbf{0.2500}
& \textbf{0.1899}\\  
\bottomrule
    \end{tabular}
    \label{tab:result_7scenes}
\end{table*} 

\begin{table*}[!t]
    \centering
    \caption{Comparison of depth estimation over the KITTI dataset~\cite{Geiger12cvpr}.} 
    \begin{tabular}{m{4em}  m{4em}   m{4.5em}   m{4.5em}  m{4em}   m{4em}   m{4em}  m{4em}   m{4em}   } 
    \toprule
  &  $\sigma<1.25$ 
  &  $\sigma<1.25^2$ 
  &  $\sigma<1.25^3$
  & abs. rel 
  & sq. rel
  & rmse  
  & rmse log
  & scale. inv \\ 
 \midrule
Eigen~\cite{Eigen14depth} 
& 67.80
& 88.79
& 96.51
& 0.1904
& 1.263
& 5.114
& 0.2758
& 0.2628
\\  
Mono~\cite{Godard17MonoDepth}  
& 86.43 
& 97.70
& \textbf{99.47}
& 0.1238
& 0.5023
& 2.8684
& 0.1644 
& 0.1635
\\  
DORN~\cite{Fu18DORN}  
& 92.62 
& \textbf{98.18}
& 99.35
& \textbf{0.0874} 
& \textbf{0.4134}
& 3.1375
& 0.1337
& 0.1233  \\  
Ours 
& \textbf{93.15} 
& 98.018
& 99.25
& 0.0998 
& 0.4732
& \textbf{2.8294}
& \textbf{0.1280}
& \textbf{0.1070} \\ 
\bottomrule
    \end{tabular}
    \label{tab:result_kitti}
\end{table*}

\subsection{Results on KITTI without GPS or IMU} 
In Table~\ref{tab:result_kitti_opt_pose}, 
we show the performance of our method on the KITTI dataset, in case
where only the IMU measurement are available (denoted as 'GT R'),
and neither IMU nor GPU are available (denoted as 'opt. pose').

\begin{table*}[!t]
    \centering
    \caption{Performance on KITTI dataset without GPS/IMU measurements} 
    \begin{tabular}{m{4em}  m{4em}   m{4.5em}   m{4.5em}  m{4em}   m{4em}   m{4em}  m{4em}   m{4em}   } 
    \toprule
  &  $\sigma<1.25$ 
  &  $\sigma<1.25^2$ 
  &  $\sigma<1.25^3$
  & abs. rel 
  & sq. rel
  & rmse  
  & rmse log
  & scale. inv \\ 
 \midrule
GT R
& 89.34
& 98.30
& 99.64
& 0.1178
& 0.4490
& 3.2042
& 0.1514
& 0.1509
\\  
opt. pose   
& 87.78
& 97.22
& 99.10
& 0.1201
& 0.5763
& 3.5157
& 0.1672 
& 0.1665 \\
\bottomrule
    \end{tabular}
    \label{tab:result_kitti_opt_pose}
\end{table*}

\section{Network structures}
In this section, we illustrate the network structures used in the pipeline. 

\subsection{D-Net}
We show the structure of the D-Net in Table.~\ref{tab:structure_DNet}.
In the paper, we set $D=64$.

\subsection{K-Net}
We show the structure of the K-Net in Table.~\ref{tab:structure_KNet}. 
In the paper, we set $D=64$.

\subsection{R-Net} 
We show the structure of the R-Net in Table.~\ref{tab:structure_RNet}. 
In the paper, we set $D=64$.

%%%%%%%%%%%%%%% 
%%%%%%%%

\begin{table*}[!ht]
    \centering
    \caption{K-Net structure. The operator expand($\cdot$) repeat the image intensity in the depth dimension} 
    \begin{tabular}{C{4em} C{23em} C{9em} C{9em}}
    \toprule
       Name  & Components & Input  & Output dimension   \\
       \midrule 
       Input 
       & concat(cost\_volume, expand($I_{\text{ref}}$))
       & 
       & $\frac{1}{4}$H $\times$ $\frac{1}{4}$ W $\times$ D $\times$ 4 \\ 
       \hline 
      conv\_0 &
      \begin{tabular}{c} 
            conv\_3d(3$\times$3, ch\_in=4, ch\_out=32), ReLU \\
            conv\_3d(3$\times$3, ch\_in=32, ch\_out=32), ReLU
       \end{tabular} 
      & Input
      & $\frac{1}{4}$H $\times$ $\frac{1}{4}$ W $\times$ D $\times$ 32
       \\ \hline
       conv\_1 &
      $\begin{bmatrix}
            \text{conv\_3d(3 $ \times$3, ch\_in=32, ch\_out=32), ReLU} \\
            \text{conv\_3d(3$\times$3, ch\_in=32, ch\_out=32) }
       \end{bmatrix} $ $\times$ 4 
      & conv\_0
      & $\frac{1}{4}$H $\times$ $\frac{1}{4}$ W $\times$ D $\times$ 32 \\ \hline
      conv\_2 & 
          \begin{tabular}{c} 
            conv\_3d(3$\times$3, ch\_in=32, ch\_out=32), ReLU \\
            conv\_3d(3$\times$3, ch\_in=32, ch\_out=1)
       \end{tabular} 
      & conv\_1
      & $\frac{1}{4}$H $\times$ $\frac{1}{4}$ W $\times$ D $\times$ 1 \\ \hline
      Output
      & Modified cost\_volume from the conv\_2 layer
      & 
      &  $\frac{1}{4}$H $\times$ $\frac{1}{4}$ W $\times$ D $\times$ 1 \\ 
    \bottomrule
    \end{tabular}
    \label{tab:structure_KNet}
\end{table*}

%%%%%%%%

\begin{table*}[!h]
    \centering
    \caption{R-Net structure} 
    \begin{tabular}{C{4em} C{23em} C{9em} C{9em}}
    \toprule
       Name  & Components & Input  & Output dimension   \\
       \midrule 
       Input 
       & cost\_volume from K-Net
       & 
       & $\frac{1}{4}$H $\times$ $\frac{1}{4}$ W $\times$ D \\ 
       \hline 
      conv\_0 &   
      \begin{tabular}{c} 
            conv\_2d(3$\times$3, ch\_in=64$+$D, ch\_out= 64$+$D), LeakyReLU \\
            conv\_2d(3$\times$3, ch\_in=64$+$D, ch\_out= 64$+$D), LeakyReLU
       \end{tabular} 
      & concat(Input, fusion in D-Net 
      )
      & $\frac{1}{4}$H $\times$ $\frac{1}{4}$ W $\times$ (64$+$D)
       \\ \hline
       trans\_conv\_0 &  
       transpose\_conv(4$\times$4, ch\_in=64$+$D, ch\_out=D, stride=2), LeakyReLU    
      &  conv\_0
      & $\frac{1}{2}$H $\times$ $\frac{1}{2}$ W $\times$ D \\ \hline
      conv\_1 &  
          \begin{tabular}{c} 
            conv\_2d(3$\times$3, ch\_in=32$+$D, ch\_out=32 $+$ D ), LeakyReLU \\
            conv\_2d(3$\times$3, ch\_in=32$+$D, ch\_out=32 $+$ D),LeakyReLU
       \end{tabular} 
      & concat(trans\_conv\_0, conv\_1 in D-Net 
      &   $\frac{1}{2}$H $\times$ $\frac{1}{2}$ W $\times$ (D$+$32) \\ \hline
      trans\_conv\_1
      & transpose\_conv(4$\times$4, ch\_in=32$+$D, ch\_out=D, stride=2 ), LeakyReLU
      &  conv\_1
      &  H $\times$ W $\times$ D \\ \hline 
      conv\_2  
      & 
      \begin{tabular}{c} 
            conv\_2d(3$\times$3, ch\_in=3$+$D, ch\_out=3$+$D ), LeakyReLU \\
            conv\_2d(3$\times$3, ch\_in=3$+$D, ch\_out=D ), LeakyReLU \\
            conv\_2d(3$\times$3, ch\_in= D, ch\_out=D )
       \end{tabular}  
      &  concat(trans\_conv\_1, $I_\text{ref}$)
      &  H $\times$ W $\times$ D \\ \hline
      Output
      & Upsampled and refined cost\_volume
      &
      &  H $\times$ W $\times$ D \\
    \bottomrule
    \end{tabular}
    \label{tab:structure_RNet}
\end{table*}

%%%%%%%
\newpage

\begin{table*}[!ht]
    \centering
    \caption{D-Net structure. The structure is taken from \cite{Chang18PSM}} 
    \begin{tabular}{C{4em} C{28em} C{4em} C{8em}}
    \toprule
       Name  & Components & Input  & Output dimension   \\
       \midrule 
       Input 
       & Input frame
       & & H $\times$ W $\times$ 3 \\ 
       \hline
       \multicolumn{4}{c}{\textbf{CNN Layers}} \\
       \hline
       conv0\_1  % conv layer
       & conv\_2d(3$\times$3, ch\_in=3, ch\_out=32, stride=2), ReLU
       &  Input 
       & $\frac{1}{2}$H $\times$ $\frac{1}{2}$ W $\times$ 32  \\ 
       \hline 
       conv0\_2  % conv layer
       & conv\_2d(3$\times$3, ch\_in=32, ch\_out=32 ), ReLU 
       &  conv0\_1
       & $\frac{1}{2}$H $\times$ $\frac{1}{2}$ W $\times$ 32\\ 
       \hline
       conv0\_3 % conv layer
       & conv\_2d(3$\times$3, ch\_in=32, ch\_out=32), ReLU 
       &  conv0\_2
       & $\frac{1}{2}$H $\times$ $\frac{1}{2}$ W $\times$ 32\\ 
       \hline
        conv1 % conv layer
       & $ \begin{bmatrix}
       \text{conv\_2d(3$\times$3, ch\_in=32, ch\_out=32), ReLU} \\
       \text{conv\_2d(3$\times$3, ch\_in=32, ch\_out=32)}
           \end{bmatrix}$ $\times$ 3
       &  conv0\_2
       & $\frac{1}{2}$H $\times$ $\frac{1}{2}$ W $\times$32 \\ 
       \hline
       conv1\_1 % conv layer
       & 
       conv\_2d(3$\times$3, ch\_in=32, ch\_out=64, stride=2), ReLU
       &  conv1
       & $\frac{1}{4}$H $\times$ $\frac{1}{4}$ W $\times$64 \\ 
       \hline
        conv2 % conv layer
       & $ \begin{bmatrix}
       \text{conv\_2d(3$\times$3, ch\_in=64, ch\_out=64), ReLU} \\
       \text{conv\_2d(3$\times$3, ch\_in=64, ch\_out=64)}
           \end{bmatrix}$ $\times$ 15
       &  conv1\_1
       & $\frac{1}{4}$H $\times$ $\frac{1}{4}$ W $\times$64 \\ 
       \hline
       conv2\_1 % conv layer
       & 
       conv\_2d(3$\times$3, ch\_in=64, ch\_out=128), ReLU
       &  conv2
       & $\frac{1}{4}$H $\times$ $\frac{1}{4}$ W $\times$128 \\ 
       \hline
       conv3 % conv layer
       & $ \begin{bmatrix}
       \text{conv\_2d(3$\times$3, ch\_in=128, ch\_out=128), ReLU} \\
       \text{conv\_2d(3$\times$3, ch\_in=128, ch\_out=128)}
           \end{bmatrix}$ $\times$ 2
       &  conv2\_1
       & $\frac{1}{4}$H $\times$ $\frac{1}{4}$ W $\times$ 128 \\ 
       \hline
        conv4 % conv layer
       & $ \begin{bmatrix}
       \text{conv\_2d(3$\times$3, ch\_in=128, ch\_out=128, dila=2), ReLU} \\
       \text{conv\_2d(3$\times$3, ch\_in=128, ch\_out=128,
       dila=2)}
           \end{bmatrix}$ $\times$ 3
       &  conv3
       & $\frac{1}{4}$H $\times$ $\frac{1}{4}$ W $\times$ 128 \\ 
       \hline
       \multicolumn{4}{c}{\textbf{Spatial Pyramid Layers}} \\
       \hline
        branch1 % spp layer
       & 
       \begin{tabular}{c}
            avg\_pool(64$\times$64,stride=64)  \\
            conv\_2d(1$\times$1, ch\_in=128, ch\_out=32), ReLU \\
            bilinear interpolation
       \end{tabular} 
       &  conv4
       & $\frac{1}{4}$H $\times$ $\frac{1}{4}$ W $\times$ 32 \\ 
       \hline
        branch2 % spp layer
       & 
       \begin{tabular}{c}
            avg\_pool(32 $\times$ 32,stride= 32)  \\
            conv\_2d(1$\times$1, ch\_in=128, ch\_out=32), ReLU \\
            bilinear interpolation
       \end{tabular} 
       &  conv4
       & $\frac{1}{4}$H $\times$ $\frac{1}{4}$ W $\times$ 32 \\ 
       \hline
        branch3 % spp layer
       & 
       \begin{tabular}{c}
            avg\_pool(16 $\times$ 16,stride= 16)  \\
            conv\_2d(1$\times$1, ch\_in=128, ch\_out=32), ReLU \\
            bilinear interpolation
       \end{tabular} 
       &  conv4
       & $\frac{1}{4}$H $\times$ $\frac{1}{4}$ W $\times$ 32 \\ 
       \hline
        branch4 % spp layer
       & 
       \begin{tabular}{c}
            avg\_pool(8 $\times$ 8,stride= 8)  \\
            conv\_2d(1$\times$1, ch\_in=128, ch\_out=32), ReLU \\
            bilinear interpolation
       \end{tabular} 
       &  conv4
       & $\frac{1}{4}$H $\times$ $\frac{1}{4}$ W $\times$ 32 \\ 
       \hline
       concat
       & concat(branch1, branch2, branch3, branch4,  conv2, conv4) 
       &
       & $\frac{1}{4}$H $\times$ $\frac{1}{4}$ W $\times$ 320
       \\
       \hline
        fusion
       & 
       \begin{tabular}{c}
            conv\_2d(3$\times$3, ch\_in=320, ch\_out=128), ReLU  \\
            conv\_2d(1$\times$1, ch\_in=128, ch\_out=64), ReLU  
       \end{tabular}
       & concat
       & $\frac{1}{4}$H $\times$ $\frac{1}{4}$ W $\times$ 64
       \\ 
       \hline 
       Output 
       &  The extracted image feature from the fusion layer
       & & $\frac{1}{4}$H $\times$ $\frac{1}{4}$ W $\times$ 64 \\  
    \bottomrule
    \end{tabular}
    \label{tab:structure_DNet}
\end{table*}

 \end{appendices}

\newpage

{\small
\bibliographystyle{ieee}
\bibliography{ref_Arxiv}
}

\end{document}